\documentclass[runningheads]{llncs}
\usepackage[T1]{fontenc}
\usepackage{graphicx}
\usepackage{textcomp}
\usepackage{xcolor}
\def\BibTeX{{\rm B\kern-.05em{\sc i\kern-.025em b}\kern-.08em
    T\kern-.1667em\lower.7ex\hbox{E}\kern-.125emX}}

\usepackage{microtype}
\usepackage{microtype}
\usepackage{makecell}
\usepackage{booktabs}
\usepackage{subcaption}
\usepackage{multirow}
\usepackage{url}

\usepackage{tikz}
\usepackage{collcell}
\usepackage{etoolbox}
\usepackage{amsmath}
\usepackage{amssymb,amsfonts,bm}

\newtoggle{inTableHeader}
\toggletrue{inTableHeader}
\newcommand*{\StartTableHeader}{\global\toggletrue{inTableHeader}}%
\newcommand*{\EndTableHeader}{\global\togglefalse{inTableHeader}}%

\DeclareMathOperator*{\softmax}{softmax}

\let\OldTabular\tabular%
\let\OldEndTabular\endtabular%
\renewenvironment{tabular}{\StartTableHeader\OldTabular}{\OldEndTabular\StartTableHeader}%

\newcommand*{\MinNumber}{0.1960}%
\newcommand*{\MidNumber}{0.56295} %
\newcommand*{\MaxNumber}{0.9299}%

\newcommand{\ApplyGradient}[1]{%
  \iftoggle{inTableHeader}{#1}{
    \ifdim #1 pt > \MidNumber pt
        \pgfmathsetmacro{\PercentColor}{max(min(100.0*(#1 - \MidNumber)/(\MaxNumber-\MidNumber),100.0),0.00)} %
        \hspace{-0.33em}\colorbox{green!\PercentColor!yellow}{#1}
    \else
        \pgfmathsetmacro{\PercentColor}{max(min(100.0*(\MidNumber - #1)/(\MidNumber-\MinNumber),100.0),0.00)} %
        \hspace{-0.33em}\colorbox{red!\PercentColor!yellow}{#1}
    \fi
  }}

\newcolumntype{R}{>{\collectcell\ApplyGradient}c<{\endcollectcell}}

\usepackage{inconsolata}
\usepackage[inline]{enumitem}

\begin{document}
\title{Towards Generalized Offensive Language Identification}
%
%
\author{Alphaeus Dmonte\inst{1} \and
Tejas Arya\inst{2} \and
Tharindu Ranasinghe\inst{3} \and
Marcos Zampieri\inst{1}}
\authorrunning{A. Dmonte et al.}
%
\institute{George Mason University, USA \and
Rochester Institute of Technology, USA \and
Lancaster University, UK \\
\email{admonte@gmu.edu}}
\maketitle              
\begin{abstract}
The prevalence of offensive content on the internet, encompassing hate speech and cyberbullying, is a pervasive issue worldwide. Consequently, it has garnered significant attention from the machine learning (ML) and natural language processing (NLP) communities. As a result, numerous systems have been developed to automatically identify potentially harmful content and to mitigate its impact. These systems can follow two approaches; \begin{enumerate*}[label=(\roman*)]
  \item Use publicly available models and application endpoints, including prompting large language models (LLMs)
  \item Annotate datasets and train ML models on them.
\end{enumerate*}
However, both approaches lack an understanding of how generalizable they are. Furthermore, the applicability of these systems is often questioned in off-domain and practical environments. This paper empirically evaluates the generalizability of offensive language detection models and datasets across a novel generalized benchmark: {\em GenOffense}. We answer three research questions on generalizability. Our findings will be useful in creating robust real-world offensive language detection systems.

\keywords{Offensive Language  \and Large Language Models \and Generalizability.}
\end{abstract}
\section{Introduction}

The presence of offensive posts on social media platforms leads to various negative consequences for users. Offensive posts have been linked to harmful outcomes such as increased suicide attempts \cite{10.1001/jamapediatrics.2015.0944,ijerph17124369} and mental health issues such as depression \cite{bannink2014cyber,bucur-etal-2021-exploratory}. To address these serious repercussions, content moderation is typically employed on online platforms. Given the overwhelming volume of posts, however, human moderators alone cannot handle the task effectively, necessitating the development of automatic systems to assist them \cite{vidgen-etal-2021-introducing,zia2022improving,zampieri2023target}. 

A highly effective method for constructing systems that can detect offensive language involves using publicly accessible application endpoints and models in an \textit{unsupervised} fashion. Notably, the development of openly accessible services such as perspective API \cite{10.1145/3534678.3539147} and models such as toxicBERT have greatly facilitated this approach. Furthermore, a more recent development involves the use of LLMs in a similar manner, employing specific prompts to identify offensive language \cite{10.1145/3543873.3587368}. The other most common method for offensive language identification is the \textit{supervised} approach, where a dataset is annotated to serve as training material for ML systems. The datasets can be annotated with different goals in mind depending on the sub-task they address, such as aggression, cyberbullying and hate speech \cite{waseem2017understanding} as well as following a more general taxonomy \cite{OLID}.

While both the \textit{unsupervised} and \textit{supervised} approaches have provided excellent results in specific offensive language detection use cases, their generalizability \cite{fortuna2021well,arango2019hate} and the ability to perform in unseen use cases \cite{SarkarChess,aggarwal2023hateproof,weerasooriya2023vicarious} have often been questioned. The ability to effectively generalize is consistently highlighted as a fundamental requirement for NLP models \cite{linzen-2020-accelerate,elangovan-etal-2021-memorization}. Particularly in a real-world application such as offensive language detection, generalization is crucial to ensure that the system exhibits robust, reliable, and fair behavior when making predictions on data that differs from their training data. However, to the best of our knowledge, no comprehensive evaluation of the generalizability of offensive language detection systems and datasets has been yet carried out. To fill this important gap in the literature, in this paper, we address the question of generalizability in offensive language identification.

Following \cite{hupkes2022state}, we define generalizability as the ability to perform consistently among different datasets. First, we construct a generalized offensive language detection benchmark; \textbf{\emph{GenOffense}}, collecting eight datasets extracted from different social media platforms and mapping them to a general offensive language detection taxonomy. We evaluate publicly available APIs and models, including LLMs in \textbf{\emph{GenOffense}}, and discuss the results. In the second part, we train various ML models on the training sets of these eight different datasets under different settings such as fully supervised, few-shot and zero-shot and evaluate the results. We answer three research questions as follows: 

\vspace{1mm}
\begin{itemize}
    \item {\bf RQ1 - Generalizability:} How well do the publicly available systems and the models trained on different datasets generalize?
\vspace{1mm}
    \item {\bf RQ2 - Dataset Size:} What is the impact of dataset size on generalizability? Does more data always result in better generalizability? 
\vspace{1mm}
    \item {\bf RQ3 - Domain Specificity:} What is the overlap and performance carryover between datasets collected from different platforms? 
\end{itemize}

\section{Related Work}

\paragraph{Offensive Language Detection} The problem of offensive language on social media has gained a lot of attention within the ML/NLP community. Researchers and organizations have developed systems to identify multiple types of offensive content such as {\em aggression}, {\em cyberbullying}, and {\em hate speech} \cite{davidson2017,ranasinghe2021mudes}. Perspective API \cite{10.1145/3534678.3539147} is one such free API that was trained on the Toxic Comment Classification dataset \cite{10.1145/3308560.3317593}. More recently, with the rise of LLMs such as GPT, researchers have used LLMs to detect and identify various forms of offensive language \cite{zampieri2023offenseval}. \cite{10.1145/3543873.3587368} utilized ChatGPT for hateful speech detection and showed that ChatGPT provides satisfactory results for certain prompts. In a different study, \cite{li2023hot} investigated the potential of using ChatGPT for annotating offensive comments and compared its results with those from crowdsourcing workers and the results show a high agreement. All these systems and APIs can be used in an \textit{unsupervised} way to detect offensive content. However, these systems can induce bias to the task depending on the data they were used to train.

As discussed in the introduction, the most common approach to detect offensive content is the \textit{supervised} approach, where the ML models are trained on annotated datasets. For this purpose, several datasets have been created for English \cite{OLID,davidson2017,mathew2020hatexplain}. The popular shared tasks such as OffensEval \cite{offenseval,zampieri-etal-2020-semeval}, HatEval \cite{basile2019semeval} and HASOC \cite{10.1145/3574318.3574326} have also contributed to creating some of these popular English datasets. Researchers have trained various ML models ranging from SVMs \cite{malmasi2017detecting} to neural transformers \cite{ranasinghe2019brums}. Recent studies have also fine-tuned transformer models on offensive language data and released domain-specific models such as HateBERT \cite{caselli2020hatebert} and fBERT \cite{sarkar2021fbert}. These supervised models have provided excellent results over several datasets.


\paragraph{Generalized Machine Learning} Good generalization, defined as the ability to successfully transfer representations, knowledge, and strategies from past experiences to new experiences, is a primary requisite for NLP/ ML models \cite{hupkes2022state}. Generalization has been widely investigated on different NLP tasks, including machine translation \cite{moisio-etal-2023-evaluating}, language modeling \cite{chronopoulou-etal-2022-efficient}, and semantic parsing \cite{jambor-bahdanau-2022-lagr} and is crucial to ensure robustness, reliability, and fairness \cite{sharma-buduru-2022-fatnet}. While the aforementioned offensive language detection methods have provided good results on the datasets they are evaluated, several studies have questioned their ability to perform on unseen use cases. \cite{SarkarChess} showed that hate speech classifiers often misclassify chess discussions as racist. \cite{weerasooriya2023vicarious} evaluate nine different offensive language detectors on political discussions and show that they have a low agreement. Furthermore, offensive language detection systems have been evaluated for geographic biases \cite{ghosh-etal-2021-detecting} and vulnerability to adversarial attacks \cite{10.1145/3270101.3270103}. Finally, \cite{fortuna2021well} tested multiple intra- and cross-dataset offensive language identification scenarios. However, the study is limited to a few datasets and models. To the best of our knowledge, no work exists on a comprehensive evaluation of the generalizability of offensive language detection systems, which we address in this research. 

\section{GenOffense: A Generalized Offensive Language Detection Benchmark}

The root cause for the lack of generalization research on offensive language detection is that no standard benchmark exists for the domain. While there are several popular datasets for offensive language identification, each of them has been annotated using different annotation guidelines and taxonomies. This, in theory, limits the possibility of combining existing datasets when training and evaluating robust offensive language identification models. To address this we construct the first Generalized Offensive Language Detection Benchmark; \emph{GenOffense}.\footnote{\url{https://github.com/TharinduDR/GeneralOffense.git}}

\begin{table*}[!ht]
\centering
\setlength{\tabcolsep}{5pt}
\scalebox{.90}{
\begin{tabular}{l|cc|cc|l|l}
\hline
& \multicolumn{2}{c|}{\bf Training} & \multicolumn{2}{c|}{\bf Testing} &  \\
\bf Dataset  & \bf Inst. & \bf OFF \% & \bf Inst.  & \bf OFF \% & \bf Data Sources & \bf Reference   \\ 
\hline
AHSD  & 19,822 & 0.83 & 4,956 & 0.82 & Twitter & \cite{davidson2017} \\ 
HASOC  & 5,604 & 0.36 & 1,401  & 0.35 & Twitter, Facebook &  \cite{mandl2020} \\ 
HatE  & 9,000 & 0.42 & 1,434 & 0.42 & Twitter &  \cite{basile2019semeval} \\ 
HateX  & 11,535 & 0.59  & 3,844  & 0.58 & Twitter, Gab &  \cite{mathew2020hatexplain}  \\ 
OHS  & 8,285 & 0.21 & 2,090 & 0.20 & Reddit & \cite{qian-etal-2019-benchmark}  \\
OLID & 13,240 & 0.33  & 860 & 0.27 & Twitter &  \cite{OLID} \\
TCC  & 12,000 & 0.09 & 2,500 & 0.10 & Wikipedia Talk & URL\textsuperscript{1} \\ 
TRAC  & 4,263 & 0.20 & 1,200 & 0.42 & Facebook, Twitter, YouTube & \cite{trac2-dataset} \\ 
\hline
\end{tabular}
}
\caption{The eight datasets used for \emph{GenOffense}, including the number of instances (Inst.) in the training and testing sets, the OFF \% in each set, the data source, and the reference.}
\label{tab:data}
\end{table*}

\subsection{\textit{GenOffense} Construction}

We use eight popular publicly available datasets containing English data summarized in Table \ref{tab:data} to construct \emph{GenOffense}. As the datasets were annotated using different guidelines and labels, following the methodology described in  \cite{ranasinghe-etal-2020-multilingual}, we map all labels to OLID level A \cite{OLID}, which is offensive (OFF) and not offensive (NOT). We choose OLID due to the flexibility provided by its general three-level hierarchical taxonomy below, where the OFF class contains all types of offensive content, from general profanity to hate speech, while the NOT class contains non-offensive examples. 

\begin{itemize}
    \item {\bf Level A:} Offensive (OFF) vs. Non-offensive (NOT).
    \item {\bf Level B:} Classification of the type of offensive (OFF) tweet - Targeted (TIN) vs. Untargeted (UNT).
    \item {\bf Level C:} Classification of the target of a targeted (TIN) tweet - Individual (IND) vs. Group (GRP) vs. Other (OTH).
\end{itemize}

\noindent In the OLID taxonomy, offensive (OFF) posts targeted (TIN) at an individual are often cyberbullying, whereas offensive (OFF) posts targeted (TIN) at a group are often hate speech.

\vspace{2mm}

\noindent \paragraph{AHSD} is one of the most popular hate speech datasets available. The dataset contains data retrieved from Twitter, which was annotated using crowdsourcing. The annotation taxonomy contains three classes: Offensive, Hate, and Neither. We conflate Offensive and Hate under a class OFF while neither class corresponds to OLID's NOT class. 

\paragraph{HASOC} is the dataset used in the HASOC shared task 2020. It contains posts retrieved from Twitter and Facebook. The upper level of the annotation taxonomy used in HASOC is hate-offensive vs non Non hate-offensive, which is the same as OLID's. This allows us to directly map hate-offensive to OLID's OFF class and non hate-offensive to NOT class. 

\paragraph{HatE} is the official dataset at SemEval-2019 Task 5 (HatEval), which focuses on hate speech against migrants and women. The first level of annotation contains two classes, hate speech or not, which can be mapped directly to OLID's OFF and NOT categories. 

\paragraph{HateX} is a dataset collected for the explainability of hate speech. It contains both token- and post-level annotation of Twitter and Gab posts. Post-level annotations have three classes: Hateful,  Offensive, and Normal. We map Hateful and Offensive classes to OFF class and Normal to NOT class.

\paragraph{OHS} is a dataset collected from Reddit with the goal of studying interventions in conversations containing hate speech. Full conversations/threads have been retrieved and annotated at the post-level as hateful or not hateful, which we map to OFF and NOT classes correspondingly. 

\paragraph{OLID} is the official dataset of the SemEval-2019 Task 6 (OffensEval) \cite{offenseval}. It contains data from Twitter annotated with a three-level hierarchical annotation which we described before. We adopt the labels in OLID level A as our classification labels. 

\paragraph{TCC} is the Toxic Comment Classification dataset. TCC was created for the Kaggle competition with the same name. The dataset contains Wikipedia comments with various classes such as toxic, obscene, insult, and threat merged in the OLID OFF class. The rest of the instances were mapped to the NOT class.

\paragraph{TRAC} is the dataset used in the TRAC shared task 2020 \cite{kumar-etal-2020-evaluating}. It focuses on aggression detection with three classes: overtly aggressive and covertly aggressive merged as OFF and non-aggressive which corresponds to the NOT class used in OLID. Finally, TRAC is the most heterogeneous dataset we used in terms of data sources containing posts from Facebook, Twitter, and YouTube. 

\subsection{\emph{GenOffense} Properties} We highlight the following generalization types that \emph{GenOffense} benchmark tests. These are shown as crucial generalization types by \cite{hupkes2022state}.

\paragraph{Platform Shift} \emph{GenOffense} benchmarks contains datasets from six different social media platforms. While most of the datasets are based on Twitter, \emph{GenOffense} has datasets that are based on other social media platforms such as Facebook and Reddit. Therefore, \emph{GenOffense} benchmark evaluates how the models can handle different platforms. 

\paragraph{Language Shift} The datasets included in \emph{GenOffense} range from 2017 to 2021. The language that was used to convey offense can be different from 2017 to 2021. Therefore, \emph{GenOffense} benchmark tests how the models can handle language shift. 

\paragraph{Task Shift} As we mentioned before, these datasets contained different tasks such as aggression detection, hate speech detection and offensive language detection. As a result, \emph{GenOffense} reflects these tasks and a model that can perform well in \emph{GenOffense} will generalize well across different sub-tasks. 

\paragraph{Topic Shift} Different datasets have been collected with different goals in mind depending on the `offensive language detection sub-task' they address. Therefore, each dataset in \emph{GenOffense} has different topics, and the models will be evaluated on how well they can handle different topics in the offensive language domain. 

\vspace{2mm}

\noindent Finally, upon acceptance of this paper, \emph{GenOffense} will be made available as an online platform where researchers can submit the model predictions and evaluate how the model generalizes over different datasets.

\section{Unsupervised Offensive Language Detection Models}
The following public models and APIs are evaluated in the test sets of \emph{GenOffense} without any training or fine-tuning. 

\begin{table}[!ht]
\setlength{\fboxsep}{0.97mm} 
\setlength{\tabcolsep}{0pt}

\begin{center}

\resizebox{\textwidth}{!}{
\begin{tabular}{p{5em} p{7em} R R R R R R R R | R} 
\toprule

&{\bf Models } &  AHSD & HASOC & HatE & HateX & OHS & OLID & TCC & TRAC & Avg \EndTableHeader \\
\midrule

\multirow{2}{*}{\bf I} & Perspective & 0.8603 & 0.6487 & 0.5340 & 0.6688 & 0.5578 & 0.7691 & 0.9228 & 0.6847 &  0.7058 \\
& ToxicBERT & 0.7430 & 0.6522 & 0.5283 & 0.6361  & 0.5416 & 0.7765 & 0.9606 & 0.6906 & 0.6911  \\
\midrule

\multirow{4}{*}{\bf II} & BERT & 0.1473 & 0.3951 & 0.4002 & 0.2986 & 0.4456 & 0.4328 & 0.3741 & 0.3961 &  0.3612 \\
& fBERT & 0.4589 & 0.3149 & 0.4075 & 0.3807  & 0.2403 & 0.3357 & 0.4178 & 0.3230 & 0.3599  \\
& HateBERT & 0.5335 & 0.4733 & 0.4968 & 0.5405  & 0.4466 & 0.4984 & 0.5945 & 0.3467 & 0.4913  \\
\midrule

\multirow{4}{*}{\bf III} & Davinci-003 & 0.8152 & 0.5909 & 0.4881 & 0.6075 & 0.4780 & 0.7401 & 0.7617 & 0.7454 &  0.6534 \\
 & Falcon-7B & 0.7406  & 0.6049 & 0.6033 & 0.6106  & 0.5291  & 0.7456  & 0.6178 & 0.7152 & 0.6458 \\
  & T0 & 0.6972  & 0.5005 & 0.4195 & 0.5631  & 0.5160  & 0.4907  & 0.6008 & 0.7126 & 0.5625 \\
   & MPT-7B & 0.5313  & 0.3571 & 0.3621 & 0.5240  & 0.2832  & 0.3703  & 0.2998 & 0.7466 &  0.4343\\

\midrule

\end{tabular}
}
\end{center}
\caption{Macro F1 score of the publicly available offensive language detection models. \textbf{Row I} shows public APIs/ models, \textbf{row II} shows the results of adapting transformers and \textbf{row III} shows the results for LLMs. The \textbf{Average} column shows the average score of all the experiments.} 
\label{tab:results_unsupervised}
\end{table}

\subsection{Methods}
\paragraph{Public APIs/ Models}

We evaluate \textbf{Perspective API} \cite{10.1145/3534678.3539147} and \textbf{ToxicBERT} \cite{davidson-etal-2019-racial}\footnote{ToxicBERT is available at \url{https://huggingface.co/unitary/toxic-bert}}.  \textbf{Perspective API} is a free API developed by Google Jigsaw, that leverages machine learning to identify toxic comments. This API was first trained using a BERT \cite{devlin2019bert} model, which is then distilled into monolingual CNN based models. The model was mainly trained on the TCC dataset, which we also included in \emph{GenOffense}. The model has six attributes, toxicity, severe toxicity, identity attack, insult, profanity, and threat. The model generates a score between 0 and 1 for each of these attributes. For each test dataset, we get all the attribute scores for each instance. If any of the attributes have a value greater than 0.5, we classify that instance as OFF, else it is classified as NOT.

We also evaluate \textbf{ToxicBERT} on \emph{GenOffense}. ToxicBERT is a BERT model trained primarily on the TCC dataset. The model is a multi-label classification model with six labels similar to Perspective API. We follow a similar approach to Perspective API to convert the ToxicBERT outputs into OFF and NOT classes. 

\paragraph{Adapting Transformers} We evaluate different general-purpose transformer models; BERT, and two domain-specific transformer models; fBERT \cite{sarkar2021fbert} and HateBERT \cite{caselli2020hatebert} on offensive language identification using an unsupervised approach. We classify a test sentence as positive or negative, where the positive label represents the NOT class and the negative represents the OFF class. We concatenate the last four hidden states returned by the model as the representative embeddings for the test sentence and the labels. We then find the cosine similarity between the representative embeddings of the labels and that of the test sentence. Finally, the sentence is assigned the label with the highest cosine similarity score.

\paragraph{Prompting LLMs} Finally, we evaluate how LLMs perform in \emph{GenOffense} benchmark, a recent trend as we discussed before. We use the following prompt to get a response from LLMs. 
\vspace{3mm}

\noindent\fbox{%
    \parbox{0.95\linewidth}{%
        \textit{Comments containing any form of non-acceptable language (profanity) or a targeted offense, which can be veiled or direct, are offensive comments. This includes insults, threats, and posts containing profane language or swear words. Comments that do not contain offense or profanity are not offensive. Is this comment offensive or not? Comment: 
    }%
}
\noindent }

\vspace{3mm}

\noindent We use several LLMs for prompting. We first use Davinci-003 through OpenAI API. Additionally, we use MPT-7B-Instruct, Falcon-7B-Instruct and T0-3B \cite{sanh2022multitask}. All of these models are available in HuggingFace\footnote{ MPT-7B-Instruct is available at \url{https://huggingface.co/mosaicml/mpt-7b-instruct}, Falcon-7B-Instruct is available at \url{https://huggingface.co/tiiuae/falcon-7b-instruct} and T0-3B is available at \url{https://huggingface.co/bigscience/T0_3B}} \cite{wolf-etal-2020-transformers}, and we use the LangChain implementation.

\subsection{Results}

The results of the aforementioned models are shown in Table \ref{tab:results_unsupervised}. Public APIs/models generally performed well on \emph{GenOffense} compared to the other two methods. However, LLMs also provide competitive results. From the LLMs, Davinci-003 performs best, closely followed by Falcon-7B. It is clear that recent LLMs produce better results on \emph{GenOffense}. Overall, Perspective API performed best on the \emph{GenOffense} benchmark. It provided the best results for six datasets out of eight and had the highest overall average. 

Most of the models show inconsistent results on the datasets. Particularly, all the models do not perform well on HatE and OHS datasets which indicates that these models do not generalize well across different tasks and platforms. 

\section{Training Offensive Language Detection Models}

In this section, we evaluate the \textit{supervised} ML models on \emph{GenOffense} benchmarks. We train the following ML models under different settings on the training sets in \emph{GenOffense} benchmark and evaluate on the test sets. 
\paragraph{LSTM}
We experiment with a bidirectional Long Short-Term-Memory (BiLSTM) model, which we adapted from the baseline in OffensEval 2019 \cite{offenseval}. The model consists of \textit{(i)} an input embedding layer with fasttext embedding \cite{bojanowski-etal-2017-enriching}, \textit{(ii)} a bidirectional LSTM layer, and \textit{(iii)} an average pooling layer of input features. The concatenation of the LSTM layer and the average pooling layer is further passed through a dense layer, whose output is ultimately passed through a \textit{softmax} to produce the final prediction. We used updatable embeddings learned by the model during training as the input.

\paragraph{Transformers} 
We also use transformers as a classification model, which have achieved state-of-the-art on a variety of offensive language identification tasks. From an input sentence, transformers compute a feature vector $\bm h\in\mathbb{R}^{d}$, upon which we build a classifier for the task. For this task, we implemented a softmax layer, i.e., the predicted probabilities are $\bm y^{(B)}=\softmax(W\bm h)$, where $W\in\mathbb{R}^{k\times d}$ is the softmax weight matrix and $k$ is the number of labels. For the experiments, we use the bert-large-cased and domain-specific fBERT \cite{sarkar2021fbert} and HateBERT \cite{caselli2020hatebert} available in HuggingFace \cite{wolf-etal-2020-transformers}.

\subsection{Model Configuration}

For LSTM, we used a Nvidia Tesla k80 to train the models. We divided the dataset into a training set and a validation set using 0.8:0.2 split. We performed \textit{early stopping} if the validation loss did not improve over 10 evaluation steps. For the LSTM model we used the same set of configurations mentioned in Table \ref{tab:lstm_parameter} in all the experiments. All the experiments were conducted for three times and the mean value is taken as the final reported result.  

\begin{table}[!ht]
\centering
\setlength{\tabcolsep}{4.5pt}
\scalebox{1.1}{
\begin{tabular}{ll}
\hline
\bf Parameter & \bf Value  \\ \hline
batch size & 64 \\
epochs & 3 \\
first dense layer units & 256 \\
learning rate & 1e-4 \\
LSTM units & 64 \\
max seq. length & 256        \\
 \hline
\end{tabular}
}
\caption{LSTM Parameter Specifications.}
\label{tab:lstm_parameter}
\end{table}

For transformers models, we used a GeForce RTX 3090 GPU to train the models. We divided the dataset into a training set and a validation set using a 0.8:0.2 split. For transformer models, we used the same set of configurations mentioned in Table \ref{tab:bert_parameter} in all the experiments.  We performed \textit{early stopping} if the validation loss did not improve over 10 evaluation steps. All the experiments were conducted three times and the mean value is taken as the final reported result.

\begin{table}[!ht]
\centering
\setlength{\tabcolsep}{4.5pt}
\scalebox{1.1}{
\begin{tabular}{ll}
\hline
\bf Parameter & \bf Value  \\ \hline
adam epsilon & 1e-8       \\
batch size & 64 \\
epochs & 3 \\
learning rate & 1e-5 \\
warmup ratio & 0.1    \\
warmup steps  & 0       \\
max grad norm & 1.0        \\
max seq. length & 256        \\
gradient accumulation steps & 1 \\
 \hline
\end{tabular}
}
\caption{BERT Parameter Specifications.}
\label{tab:bert_parameter}
\end{table}

\renewcommand{\arraystretch}{1.2}
\begin{table}[]
\setlength{\fboxsep}{0.97mm} 
\setlength{\tabcolsep}{0pt}


\resizebox{\textwidth}{!}{
\begin{tabular}{p{6em} l R R R R R R R R | R} 
\toprule

&{\bf \makecell{Train \\ Dataset(s)} } &  AHSD & HASOC & HatE & HateX & OHS & OLID & TCC & TRAC & Avg \EndTableHeader \\
\toprule

\multirow{11}{*}{\bf LSTM} & AHSD & 0.8872 & 0.5465 & 0.3735 & 0.4903 & 0.3757 & 0.4005 & 0.4598 & 0.5809 &  0.5143\\
& HASOC & 0.4336 & 0.6539 & 0.5388 & 0.5339  & 0.5503 & 0.5832 & 0.5756 & 0.4056 & 0.5343 \\
& HatEval & 0.6605  & 0.5200 & 0.5825 & 0.5266 & 0.5479 & 0.5413 & 0.5212 & 0.4991 & 0.5498 \\
& HateX & 0.5531 & 0.4623 & 0.3976 & 0.7091 & 0.4943 & 0.5193 & 0.3710 & 0.4710 & 0.4927 \\
& OHS & 0.1487 & 0.3936 & 0.5234 & 0.5309 & 0.6984 & 0.4670 & 0.2604 & 0.8117 &  0.4793 \\
& OLID & 0.6391 & 0.6224 & 0.5283 & 0.5477 & 0.5636 & 0.7124 & 0.6366 & 0.7473 &  0.6247\\
& TCC & 0.5432 & 0.4756 & 0.5581 & 0.5497 & 0.5841 & 0.5711 & 0.7930 & 0.5587 & 0.5791 \\
& TRAC & 0.1800 & 0.4058 & 0.5105 & 0.4868 &  0.5376 & 0.5457  & 0.5460 & 0.6853 &  0.4872 \\
\cmidrule{2-11}
& All & 0.8689 & 0.6134 & 0.4849 & 0.6775 & 0.6236 & 0.6754 & 0.6537 & 0.7490 &  0.6681 \\
& All-1 & 0.8675 & 0.5745 & 0.4539 & 0.5842 & 0.4957 & 0.5569 & 0.6491 & 0.6231 &  0.6006 \\
\midrule

\multirow{11}{*}{\bf BERT} & AHSD & 0.9268 & 0.6300 & 0.5279 & 0.5867 & 0.5179 & 0.6991 & 0.8188 & 0.6278 &  0.6657 \\
& HASOC & 0.6203 & 0.7585 & 0.5850 & 0.5550  & 0.5798 & 0.4925 & 0.6541 & 0.5495 &  0.5993 \\
& HatEval & 0.6122 & 0.4418 & 0.5880 & 0.4966 & 0.5795 & 0.5795 & 0.6240 &  0.6884 &  0.6012 \\
& HateX & 0.5690 & 0.6049 & 0.6322 & 0.7829 & 0.6167 & 0.5049 & 0.7214 & 0.5382 & 0.6212 \\
& OHS & 0.1960 & 0.4100 & 0.4567 & 0.3875 & 0.7745 & 0.4225 & 0.5048 & 0.3920 & 0.4430 \\
& OLID & 0.6857 & 0.6366 & 0.5296 & 0.6206 & 0.5725 & 0.8074 & 0.8451 & 0.7402 & 0.6797 \\
& TCC & 0.7210 & 0.6448 & 0.5241 & 0.6297  & 0.5677 & 0.7453  & 0.8805 & 0.6678 & 0.6726 \\
& TRAC & 0.6225 & 0.6260 & 0.5757 & 0.6122 & 0.5579 & 0.6916  & 0.7692 & 0.8596 &  0.6643 \\
\cmidrule{2-11}
& All & 0.9257 & 0.7506 & 0.7412 & 0.7718 & 0.7263 & 0.7449  & 0.8578 & 0.7793 &  0.7872 \\
& All-1 &  0.3805 & 0.5346 & 0.5557 & 0.5771 & 0.5652 & 0.6680 & 0.7829 & 0.6529 &  0.5896 \\



\midrule

\multirow{11}{*}{\bf HateBERT} & AHSD & 0.9299 & 0.6248 & 0.5367 & 0.6051 & 0.5311 & 0.6425 & 0.7313 & 0.5813 &  0.6478 \\
& HASOC & 0.5704 & 0.6529 & 0.5852 & 0.5873  & 0.5563 & 0.6666 & 0.6101 & 0.7247 & 0.6192  \\
& HatEval & 0.7033 & 0.4974 & 0.4748 & 0.5852 & 0.5392 & 0.4814 & 0.6336 &  0.5421 &  0.5571 \\
& HateX & 0.5276 & 0.5954 & 0.5765 & 0.7724 & 0.5805 & 0.4981 & 0.6547 & 0.5609 & 0.5958 \\
& OHS & 0.2149 & 0.4024 & 0.3785 & 0.3102 & 0.7591 & 0.4189 & 0.4982 & 0.3651 & 0.4184 \\
& OLID & 0.7610 & 0.6239 & 0.5465 & 0.5971 & 0.4855 & 0.7811 & 0.7822 & 0.6317 & 0.6511 \\
& TCC & 0.7885 & 0.6286 & 0.5376 & 0.6386  & 0.5502 & 0.7107  & 0.8408 & 0.6493 & 0.6680 \\
& TRAC & 0.2597 & 0.5083 & 0.5164 & 0.5614 & 0.5715 & 0.5838  & 0.6392 & 0.8239 & 0.5580  \\
\cmidrule{2-11}
& All & 0.9174 & 0.6180 & 0.6076 & 0.7803 & 0.7009 & 0.7278 & 0.7955 & 0.6789 & 0.7283  \\
& All-1 & 0.4844  & 0.5487 & 0.5760 & 0.6073 & 0.5751 & 0.5257 & 0.7861 & 0.6625 & 0.5957  \\
\midrule

\multirow{11}{*}{\bf fBERT} & AHSD & 0.9241 & 0.6365 & 0.5318 & 0.6246 & 0.5096 & 0.6918 & 0.8032 & 0.5482 & 0.6587  \\
& HASOC & 0.6912 & 0.6753 & 0.5386 & 0.6343  & 0.5510 & 0.7778 & 0.8226 & 0.7443 & 0.6794  \\
& HatEval & 0.6810 & 0.5332 & 0.4917 & 0.5724 & 0.5693 & 0.5599 & 0.6893 &  0.6714 &  0.5960 \\
& HateX & 0.5276 & 0.5954 & 0.6263 & 0.7840 & 0.5784 & 0.5252 & 0.7156 & 0.5991 & 0.6189 \\
& OHS & 0.1615 & 0.3935 & 0.4649 & 0.5720 & 0.7558 & 0.5572 & 0.6905 & 0.5382 & 0.5167 \\
& OLID & 0.7239 & 0.6572 & 0.5474 & 0.6217 & 0.5217 & 0.7838 & 0.8234 & 0.7524 & 0.6789 \\
& TCC & 0.7497 & 0.6545 & 0.5243 & 0.6303  & 0.5452 & 0.7458  & 0.8486 & 0.6753 & 0.6717 \\
& TRAC & 0.5757 & 0.5975 & 0.5565 & 0.6277 & 0.5389 & 0.7049  & 0.8290 & 0.8416 & 0.6589  \\
\cmidrule{2-11}
& All & 0.9201 & 0.6338 & 0.5727 & 0.7768 & 0.7102 & 0.7350 & 0.8389 & 0.7677 & 0.7444  \\
& All-1 & 0.3516  &  0.5590 & 0.5588 & 0.6369 & 0.5685 & 0.5854 & 0.7889 & 0.6601 &  0.5887 \\

\bottomrule
\end{tabular}
}
\caption{Macro F1 score of the offensive language detection models. The \textbf{Training Dataset(s)} shows the training dataset while the subsequent columns show the results for each test set. The \textbf{Average} column shows the average score of all the experiments.} 
\label{tab:results_domain}
\end{table}

\subsection{Results}

We use multiple strategies to answer the three \textbf{RQ}s considering generalizability with respect to training and testing data. 

We address training set variation by training the three models in the following settings:

\paragraph{1 to 1} We train a separate machine learning model on each of the eight training sets. We then evaluate the trained model on each of the eight test sets in isolation. 
\paragraph{All -1} We concatenate all training sets except one and train a single machine learning model. We then evaluate the model on the test set of that particular dataset that was left out. 
\paragraph{All} We concatenate the training sets of all the datasets and trained a single machine learning model. We then evaluate the model on each testing set of all eight datasets in \emph{GenOffense}.
\paragraph{Few to 1} We also perform progress tests. We randomly selected 1000, 2000, 3000 etc. instances from each of the eight training sets and train separate machine learning models. We then evaluate the trained model on each of the eight test sets in isolation.  

\begin{figure}[!ht]
\centering
\includegraphics[scale=0.6]{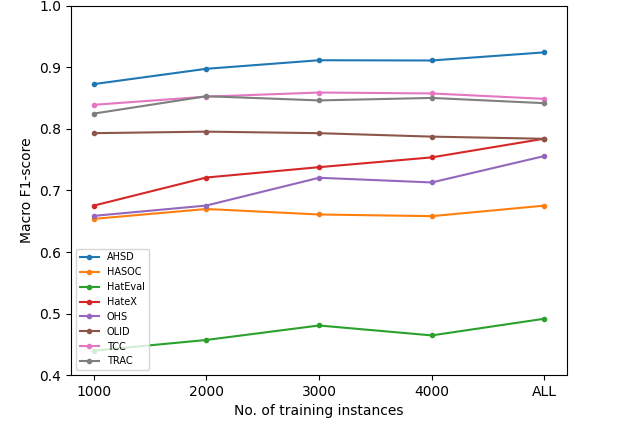}
\caption{Few-shot Learning Results for BERT}
\label{fig:fbert}
\end{figure}

We present the results of the aforementioned strategies in Table \ref{tab:results_domain} and Figure \ref{fig:fbert} in terms of Macro F1 score. The transformer models outperform the LSTM for all tested dataset combinations. This is in line with the findings of popular competitions such as HatEval and OffensEval. However, domain-specific models such as fBERT and HateBERT did not outperform BERT in average scores of \emph{GenOffense}. This can be because both of these models are fine-tuned on platform-specific data. Unsurprisingly the {\bf all} strategy, achieves the best results in all four classifiers. However, few-shot results in Figure \ref{fig:fbert} suggest and more training instances do not improve the average Macro F1 score of \emph{GenOffense}. Furthermore, the {\bf all -1} strategy was outperformed by many of the individual datasets suggesting that simply using a large dataset does not always result in better generalizability. 

In terms of the individual dataset performance, models trained on OLID yielded the highest generalization followed by TCC. This is due to the general nature of these two datasets covering multiple types of offensive content rather than focusing on a particular type of offensive content (e.g. hate speech). AHSD also provided good generalization, likely due to the presence of both hate speech and general offensive language in the dataset. On the other hand, models trained on OHS yielded the worst performance. This can be explained by the platform-specificity of the dataset, as OHS is the only Reddit dataset in this collection. 

\subsection{Test Set Combination} 

We also look at the performance of the models on a single test set combining all individual test sets in \emph{GenOffense}. We use a separate BERT model on each of the eight training sets and tested them on the concatenated test set. We present the results obtained on the consolidated test set in terms of Macro and Weighted F1 Table \ref{tab:all}. 

\begin{table}[!ht]
\centering
\setlength{\tabcolsep}{4.5pt}
\resizebox{0.6\textwidth}{!}{
\begin{tabular}{lcc}
\toprule
\bf Train Dataset  & \bf Macro F1 & \bf Weighted F1 \\ 
\midrule
AHSD & 0.7348   &  0.7348    \\
HASOC & 0.6722   & 0.6743      \\
HatE      &  0.6210  &  0.6239    \\
HateX      & 0.6879   &   0.6899  \\
OHS & 0.4247 &	0.4348 \\
OLID & \bf 0.7543 &	\bf 0.7551 \\
TCC &  0.7064 &	0.7060 \\
TRAC & 0.6467 & 0.6492	 \\
\bottomrule
\end{tabular}
}
\caption{BERT results for the combined test set in terms of Macro F1 and Weighted F1. Best results in bold.}
\label{tab:all}
\end{table}

\noindent The results indicate that models trained on OLID offers the best performance on the combined test set, followed by AHSD, and TCC while OHS delivers the lowest performance by a very large margin. This is in line with the results obtained using individual test sets. 

\section{Conclusion}

This paper introduced the first generalization benchmark for offensive language detection; \emph{GenOffense}. We also presented a comprehensive evaluation of the generalizability of different computational models, including recently released LLMs. We hope that our findings motivate the community to further explore the question of generalizability as argued by other recent studies \cite{fortuna2020toxic,fortuna2021well}.

We revisit the research questions posed in the introduction: 
 
 \begin{itemize}
 
\item {\bf RQ1 - Generalizability:} Despite being popular, LLMs did not perform well in the \emph{GenOffense} benchmark. APIs, such as Perspective, showed better generalizability. In the supervised setting, models trained on OLID, AHSD, and TCC provided the best generalizability to other datasets. This can be explained by their focus on general offensive language (in the case of OLID and TCC) and the presence of both hate speech and general offensive (in the case of AHSD), which is reflected in their annotation models. More specific datasets, such as HatEval, which focuses on women and migrants, displayed lower results. Finally, OHS, the only Reddit dataset, achieved the lowest performance, suggesting that the domain has a substantial impact on performance (see {\bf RQ3}).  

\vspace{1mm}
\item {\bf RQ2 - Dataset Size:} We observed that more data does not always result in better generalizability. The few-shot experiments showed that adding more training instances did not provide better generalizability. Even though the "All" strategy achieved the best performance for all datasets, the "All-1" strategy achieved performance lower than most datasets in isolation. Therefore we have not found a direct correlation between generalizability and training dataset size in our experiments. The question of dataset size requires further investigation.

\vspace{1mm}
\item {\bf RQ3 - Domain Specificity:} Models trained on OHS, the only Reddit dataset in the collection, achieved the lowest performance of all datasets, suggesting that the domain plays an important role in generalizability. OHS is not the smallest dataset tested in our experiments, therefore we believe that the low performance is due to the specificity of their source material (Reddit) rather than its size. We would like to further investigate this by running more dataset ablation experiments. 

\end{itemize}

\noindent In future work, we would like to extend \emph{GenOffense} benchmark to adversarial test sets using popular augmentation techniques such as random insertion and random deletion. This will provide the opportunity for the researchers to explore probing in offensive language detection models. We believe this would provide us with even more insights into the generalizability of the datasets and the robustness of the models. Finally, we would like to extend \emph{GenOffense} to support multilingual offensive language datasets and 
replicate these experiments for different languages. Such multilingual benchmarks will be useful for many real-world applications.

\section*{Acknowledgements}

We would like to thank the anonymous reviewers for their positive and valuable feedback. We further thank the creators of the datasets used in this paper for making the datasets publicly available for our research. 

The experiments in this paper were conducted on the High End Computing (HEC) Cluster at Lancaster University, which is funded through a combination of central funding and contributions from individual research grants. The experiments were designed in UCREL-HEX\cite{UcrelHex}, which is a collection of GPU equipped hosts at the School of Computing and Communications, Lancaster University.

Marcos Zampieri is partially supported by a grant from the Virginia Commonwealth Cyber Initiative (CCI) award number N-4Q24-009 . 




%
%
%
\bibliographystyle{splncs04}
\bibliography{custom}

\end{document}